# DSL: Discriminative Subgraph Learning via Sparse Self-Representation

Lin Zhang[*]   Petko Bogdanov[†]

**Abstract**

The goal in network state prediction (NSP) is to classify the global label associated with features embedded in a graph. This graph structure encoding feature relationships is the key distinctive aspect of NSP compared to classical supervised learning. NSP arises in various applications: gene expression samples embedded in a protein-protein interaction (PPI) network, temporal snapshots of infrastructure or sensor networks, and fMRI coherence network samples from multiple subjects to name a few. Instances from these domains are typically "wide" (more features than samples), and thus, feature sub-selection is required for robust generalizable prediction. How to best employ the network structure in order to learn succinct connected subgraphs encompassing the most discriminative features becomes a central challenge. Prior work employs connected subgraph growth and sampling or graph smoothing within optimization frameworks, resulting in either large variance of quality or weak control over the connectivity of selected subgraphs.

In this work we propose an optimization framework for discriminative subgraph learning (DSL), which simultaneously enforces (i) sparsity, (ii) connectivity and (iii) high discriminative power of the resulting subgraphs of features. Our optimization algorithm is a single-step solution for the NSP and associated feature selection problem. It is rooted in the rich literature on maximal-margin optimization, spectral graph methods and sparse subspace self-representation. DSL simultaneously ensures solution interpretability and superior predictive power (up to 16% improvement in challenging instances compared to baselines), with execution times up to an hour for large instances.

## 1 Introduction

Global network state prediction (NSP) [16, 23, 8, 7] is a supervised learning problem in which features are embedded in a network as node/edge weights. The basic premise in this setting is that the global state of the network is determined by local network processes which modify connected feature values in a predictable manner. *Given a set of network samples over the same nodes and similar or identical interconnecting structure, how to select select connected subgraphs to accurately predict the global network states?*

The NSP problem arises in multiple application domains: phenotype prediction based on gene expression within a protein interaction network [16], learning rate prediction based on functional MRI scans [3], congestion/normal regime prediction in communication networks, prediction of global phenomena in spatial sensor network samples and others. Common to all the above settings is the importance of network locality in selecting predictive features for the global state. In addition, datasets fitting this setting are typically "wide": involving many more features than labeled instances. Hence, it becomes imperative to learn robust and general predictors of the global state when only a subset of the features are considered, a problem commonly referred to as feature selection [5]. The distinctive characteristic of NSP is that the network structure can be exploited to detect robust and interpretable feature subsets.

Intuitively, a good solution for the problem should identify a small number of features *(i) sparsity*, forming connected subgraphs *(ii) connectivity*, whose feature values accurately predict the global state *(iii) discriminative power*. Satisfying all three design principles simultaneously is a challenging task, hence, prior work typically prioritizes a subset of them. Some methods enforce connectivity by directly growing [16] or sampling [23] connected subgraphs. Such approaches suffer limited prediction quality and/or instability due to the local exploration of the exponential space of connected subgraphs. Other approaches enforce the design principles within optimization frameworks [8, 7]. Due to the inherent complexity of simultaneous optimization of all three, these methods partition the principles in independent steps resulting in sub-par performance.

An illustration of the above phenomenon is demonstrated in Fig. 1a by superimposing the subgraphs selected by (i) L1DSL (one of the methods proposed in this paper), (ii) the state-of-the-art optimization approach DIPS [8] and a ground truth (GT) subgraph injected in a Synthetic dataset. When constrained to select a fixed-size subgraph, our proposed method L1DSL recovers the connected GT subgraph, while DIPS recovers the GT only partially due to its two-step independent enforcement of the three design principles. Moreover our method consistently recovers 90% of the GT nodes for increasing noise added to non-GT feature values (Fig. 1b), while the baseline's accuracy quickly degrades due to its susceptibility to noise.

---

[*]University at Albany—SUNY, lzhang22@albany.edu
[†]University at Albany—SUNY, pbogdanov@albany.edu

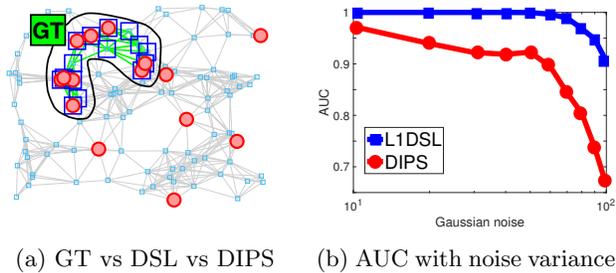

(a) GT vs DSL vs DIPS  (b) AUC with noise variance

Figure 1: (a): Comparison of the subgraphs detected by DSL (blue squares) and DIPS [8] (red circles) superimposed with a ground truth GT subgraph in a synthetic network dataset. (b) AUC of GT subgraph recovery for L1DSL and DIPS for increasing Gaussian noise added to non-GT feature values.

We address drawback of past work on NSP by proposing *DSL* (pronounced DieSeL): an optimization framework for NSP and the associated feature selection problem. We enforce the three design principles discussed above in a unified objective function and propose an algorithm for its optimization. We enforce sparsity by a self-representation objective designed to consistently select a sparse subset of features in all training network samples. Connectivity and discriminative power are enforced by appropriate regularization inspired by spectral graph methods and subspace max-margin optimization. We construct a solver for the objective function and study its utility and effectiveness on synthetic and real-world datasets from multiple domains, demonstrating its superior quality compared to several baselines.

Our main contributions in this work are as follows:
**1. Novelty:** We combine all three design principles for NSP: sparsity, connectivity and discriminative power in a novel unified optimization framework, called DSL.
**2. Quality:** DSL consistently outperforms state-of-the-art baselines on real-world and synthetic datasets for both recovery of ground truth subgraphs and in classification accuracy ($7\% - 16\%$ improvement compared to baselines) employing small-size selected subgraphs.
**3. Interpretability and wide applicability:** DSL discovers interpretable feature subgraphs: known genes associated with liver metastasis in PPI networks and natural "corridor" patterns of bike commute behavior distinguishing between workdays and weekends.

## 2 Related work

Early existing methods for NSP focus on direct exploration of the space of connected subgraphs [16, 23]. NGF [16] explores the structure of PPI networks employing the random forest classifier iteratively fitted to growing connected structural subgraphs. MINDS [23] adopts a similar tree construction, while seeking to improve the running time and quality by a Markov Chain Monte Carlo (MCMC) sampling scheme in the subgraph space. Both methods suffer from limited quality and high running time due to the need to explore a large space of connected subgraphs.

An alternative family of approaches for NSP were recently proposed following an optimization strategy [7, 8]. DIPS [8] is the state-of-art approach, which introduces a two-stage solution to learn subgraphs: discriminative subspace learning followed by matrix approximation. This method avoids the search in the exponential space of candidate subgraphs, thus, addressing major drawbacks of NGF and MINDS. However, its $\ell_1$-norm based node selection mechanism is sensitive to noise and outliers and in addition subgraphs selection is performed in two independent steps, limiting the quality of obtained solutions.

A different subspace learning problem is Sparse Subspace Clustering (SSC) [13, 19], where the goal it to approximate unlabeled data by selecting a few feature comprising a subspace. This problem, however, is unsupervised and does not consider a network structure among features.

There is also related work on support vector machines (SVM) [6, 27]. SVM as a classification scheme finds an optimal separating hyperplane between classes. SVMs have been combined with subspace learning approaches such as matrix factorization [24, 10]. Our work is different from the above in that we consider a graph structure among features and employ self-representation as opposed to matrix factorization. We employ both the $\ell_2$-SVM [6] and $\ell_1$-SVM [27] models as regularizers in our objective. The L1-norm SVM[27] learns to ignore redundant features (sparsity), thus, allowing for automatic feature selection, which makes it often a better choice in very high dimensional data.

## 3 Notation and preliminaries

The global network state prediction can be viewed as a generalization of the classical supervised learning problem, where the knowledge of the network structure can be employed to improve the classification as well as provide explanation for the selected features. We first introduce the notation and preliminaries employed in our problem formulation and solution.

The input to the problem is a set of network samples (or graph signals). A *network sample* is a triple $S_i = (\mathcal{V}_i, E_i, X_i)$, where $\mathcal{V}_i = v_1, v_2, ..., v_m$ is a set of nodes, $E_i \subseteq \mathcal{V}_i \times \mathcal{V}_i$ is a set of undirected edges, and $X_i$ is a function labeling each node with a real number. The function $X_i$ can be thought of as a graph signal over the nodes of the sample.

Let $\mathcal{DS} = \{(S_1, y_1), (S_2, y_2), ..., (S_n, y_n)\}$ be a network dataset that consists of $n$ network samples annotated by corresponding discrete global states (or labels) $y_i$. Similar to Dang et al. [8], we adopt a summary graph structure $S = (\mathcal{V}, E, W)$ to represent all $S_i \in \mathcal{DS}$, where $\mathcal{V} = \mathcal{V}_1 \cup \mathcal{V}_2 ... \cup \mathcal{V}_n$, $E \subseteq \mathcal{V} \times \mathcal{V}$ and $E_i \subseteq E, \forall E_i$. Each edge $E(p, q) \in E$ is associated with a positive weight $W_{pq}$ defined as the fraction of network samples containing that edge in their structure, i.e., $W_{pq} = n^{-1} \times \sum_i E_i(p, q)$ with $E_i(p, q) = 1$ if $v_p$ connects $v_q$ in $S_i$. The combinatorial Laplacian matrix $\mathbf{L}$ associated with the aggregate network $S$ is defined as $\mathbf{L} = D - W$, where $D$ is the diagonal matrix of weighted node degrees with elements $D_{pp} = \sum_q W_{pq}$.

Arranging the node values of all networks samples $X_i$ in the columns of a matrix, we obtain the data matrix $\mathbf{X} \in \mathbb{R}^{m \times n}$, where $n$ is the total number samples and $m$ is the total number of nodes in the network, also referred to as the feature dimension. In our formulation will enforce the selection of connected features (rows of $\mathbf{X}$). In the sparse subspace clustering literature [13] such selection is enforced through the product $\mathbf{X}^T \mathbf{\Phi}$, where $\mathbf{\Phi}$ is a feature selection matrix with zero elements on the diagonal diag($\mathbf{\Phi}) = \mathbf{0}$ to avoid individual columns being represented solely by themselves [13].

## 4 DSL: discriminative subgraph learning

Our goal is to simultaneously select connected subgraphs which are predictive of the global state. We formalize the problem as an optimization which linearly combines (i) selection of subgraphs, (ii) connectivity and (iii) discriminative power of the selection on the training network samples.

**Subspace selection.** We enforce selection of a representative subset of features by minimizing the reconstruction error for the data matrix $\mathbf{X}$ via its subspace representation through an unknown feature selection matrix $\mathbf{\Phi}$: $\left\|\mathbf{X}^T - \mathbf{X}^T \mathbf{\Phi}\right\|_F^2$, where the reconstruction error is quantified in terms of the Frobenius norm of the residual matrix. To control how many features we select, we need to control the sparsity of the selection matrix $\mathbf{\Phi}$. A widely adopted approach is to add an $\ell_1$-norm regularizer $\left\|\mathbf{\Phi}\right\|_1$, however, this choice would not enforce that the same node feature is selected across network samples. Intuitively we would like rows of $\mathbf{\Phi}$ to contain only high values (corresponding node is selected) or only values close to 0. To this end, we adopt the $\ell_{2,1}$-norm for the selection matrix defined as $\|\mathbf{\Phi}\|_{2,1} = \sum_i \sqrt{\sum_j \mathbf{\Phi}_{ij}^2} = \sum_i \|\mathbf{\Phi}_i\|_2$, where $\mathbf{\Phi}_i$ is the $i$-th row of $\mathbf{\Phi}$.

**Subgraph connectivity.** Our second goal is to ensure that our selection of nodes encoded in $\mathbf{\Phi}$ is also smooth (connected) with respect to the summary graph structure $S$ interconnecting features. We achieve this by a regularizer involving the trace of the following quadratic form of the Laplacian matrix: $\text{tr}\left(\mathbf{\Phi}^T \mathbf{L} \mathbf{\Phi}\right)$. Each diagonal element in the product is of the form:

$$(4.1) \quad (\mathbf{\Phi}^T \mathbf{L} \mathbf{\Phi})_{k,k} = \mathbf{\Phi}_k^T \mathbf{L} \mathbf{\Phi}_k = \sum_{(i,j) \in E} w_{ij} (\mathbf{\Phi}_{ik} - \mathbf{\Phi}_{jk})^2,$$

where $\mathbf{\Phi}_k$ is the $k$-th column in $\mathbf{\Phi}$, i.e. the selector vector for the $k$-th instance. Intuitively, this criterion penalizes for selection of non-neighbor features in each network sample.

**Discriminative power.** Our third goal is to ensure that the selected subgraphs are disriminative, in other words the included features should be able to correctly separate networks instances with different global states. We employ a loss function inspired by maximal margin optimization in SVM. Intuitively, feature values in selected subgraphs should render different class instances on opposite sides of a separation hyperplane $(\mathbf{w}, b)$, such that the margin defined by support vectors is maximized. We further allow for soft margin to avoid overfitting.

**DSL objective.** Incorporating the above principles into a single objective, we obtain the following optimization for discriminative subgraph learning:

$$\operatorname*{argmin}_{\mathbf{\Phi}, \mathbf{w}, b} \left\|\mathbf{X}^T - \mathbf{X}^T \mathbf{\Phi}\right\|_F^2 + \lambda_1 \|\mathbf{\Phi}\|_{2,1} + \lambda_2 \text{tr}\left(\mathbf{\Phi}^T \mathbf{L} \mathbf{\Phi}\right)$$

$$+ \pi \left\{ \|\mathbf{w}\|_f + C \sum_i^n \ell\left(y_i, \mathbf{w}^T \hat{\mathbf{x}}_i + b\right) \right\}, \text{ s.t diag}(\mathbf{\Phi}) = \mathbf{0}$$

The first two terms in the objective reflect the subspace learning, the third term incorporates smoothness with respect to the graph structure, while the last term captures the soft margin maximization. The function $\ell\left(y_i, \mathbf{w}^T \hat{\mathbf{x}}_i + b\right)$ is the hinge loss function, in which $\mathbf{w}$ is the normal vector to the hyperplane, $b$ is an offset term, and $C$ is the soft-margin control parameter. Each regularization has a corresponding balance parameter, namely $\lambda_1$ and $\lambda_2$ and $\pi$, controlling the importance of sparsity, graph smoothness and discriminative power respectively.

The norm $f$ on the vector $\mathbf{w}$ orthogonal to the separation hyperplane $\|\mathbf{w}\|_f$ is either $f = 1$ or $f = 2$, giving rise to two flavors of DSL: L1DSL and L2DSL respectively. In addition, the notation $\hat{\mathbf{x}}_i = \mathbf{\Phi}^T \mathbf{x}_i$ denotes the projection of the $i$-th sample value onto the selection matrix $\mathbf{\Phi}$. Intuitively, we are penalizing misclassification based on the sub-selection of the features through $\mathbf{\Phi}$ as opposed to when considering all features. Note that we also add the 0 constraint for diagonal entries of the selection matrix $\mathbf{\Phi}$, a typical constraint in sparse subspace learning to prevent the selection matrix from representing each feature by itself as opposed to as linear combination of other features.

By design our DSL objective can be viewed as a natural generalization of sparse subspace clustering, spectral graph partitioning and maximal margin learning. We optimize all those objectives simultaneously to learn discriminative connected subgraphs.

## 5 Optimization for DSL: Learning Algorithm

The optimization in the DSL objective is with respect to two sets of parameters: the selection matrix $\mathbf{\Phi}$, and the orthogonal vector to the separation hyperplane including offset $(\mathbf{w}, b)$. Since the hinge loss, $\ell_{2,1}$-norm and trace norm are not smooth, it's hard to develop an optimization on them simultaneously as Nesterov method [20]. Hence we design an alternating minimization method to optimize $\mathbf{\Phi}$ and $(\mathbf{w}, b)$. We next outline the optimization of the corresponding subproblems and list the steps of the overall algorithm. During the iteration, we first ignore the constraint of zero on the diagonal and add it to the result after the iteration.

**Updates for $\mathbf{\Phi}$.** When the separating hyperplane and offset $(\mathbf{w}, b)$ are fixed, the optimization simplifies to:

$$(5.2) \quad \underset{\mathbf{\Phi}}{\operatorname{argmin}} \left\| \mathbf{X}^T - \mathbf{X}^T \mathbf{\Phi} \right\|_F^2 + \lambda_1 \left\| \mathbf{\Phi} \right\|_{2,1} + \lambda_2 \operatorname{tr}\left( \mathbf{\Phi}^T \mathbf{L} \mathbf{\Phi} \right) + C^* \sum_i^n \ell\left( y_i, \mathbf{w}^T \mathbf{\Phi}^T \mathbf{x}_i + b \right),$$

where $C^* = \pi C$. Optimizing directly with the non-smooth hinge loss function is challenging. Hence, in line with the optimization literature on SVMs, we introduce slack variables for each instance $\xi_i$, separating the non-smooth hinge loss in constraints:

$$(5.3) \quad \underset{\mathbf{\Phi}, \xi}{\operatorname{argmin}} \left\| \mathbf{X}^T - \mathbf{X}^T \mathbf{\Phi} \right\|_F^2 + \lambda_1 \left\| \mathbf{\Phi} \right\|_{2,1} + \lambda_2 \operatorname{tr}\left( \mathbf{\Phi}^T \mathbf{L} \mathbf{\Phi} \right) + C^* \sum_i^n \xi_i$$

s.t $y_i \left( \mathbf{w}^T \mathbf{\Phi}^T \mathbf{x}_i + b \right) \geq 1 - \xi_i; \; \xi_i \geq 0$

To solve the problem in Eq. 5.3, we construct the corresponding Lagrangian function and derive a closed-form update. The Lagrangian has the following form:

$$(5.4) \quad \begin{aligned} L\left(\mathbf{\Phi}, \xi, \alpha, \gamma\right) &= \left\| \mathbf{X}^T - \mathbf{X}^T \mathbf{\Phi} \right\|_F^2 + \lambda_1 \left\| \mathbf{\Phi} \right\|_{2,1} \\ &+ \lambda_2 \operatorname{tr}\left( \mathbf{\Phi}^T \mathbf{L} \mathbf{\Phi} \right) + C^* \sum_i^n \xi_i - \sum_i^n \gamma_i \xi_i \\ &- \sum_i^n \alpha_i \left[ y_i \left( \mathbf{w}^T \mathbf{\Phi}^T \mathbf{x}_i + b \right) - 1 + \xi_i \right], \end{aligned}$$

where $\boldsymbol{\alpha}$ and $\boldsymbol{\gamma}$ are vectors of Lagrangian multipliers of length $\mathcal{DS}$. Setting the gradient $\nabla_{\mathbf{\Phi}} L = 0$, we obtain:

$$\mathbf{\Phi} = \frac{\left( \mathbf{X} \mathbf{X}^T + \lambda_1 \mathbf{D} + \lambda_2 \mathbf{L} \right)^{-1}}{2} \left( \sum_i \alpha_i y_i \mathbf{x}_i \mathbf{w}^T + 2 \mathbf{X} \mathbf{X}^T \right),$$

where $\mathbf{D}$ is a diagonal matrix with elements $D_{ii} = \left( 2 \left\| \mathbf{\Phi}_i \right\|_2 \right)^{-1}$ when $\left\| \mathbf{\Phi}_i \right\|_2 \neq 0$, and $D_{ii} = 0$ otherwise. Similarly, solving $\nabla_{\xi} L = 0$, we get:

$$(5.5) \quad \boldsymbol{\gamma} = C^* \mathbb{1} - \boldsymbol{\alpha},$$

where $\mathbb{1}$ is a vector of ones of size $|\mathcal{DS}|$. By substituting the optimal values of $\mathbf{\Phi}$ and $\boldsymbol{\gamma}$ back in the Lagrangian (Eq. 5.4), and after some simplifying variable substitutions we obtain the following dual Lagrangian function:

$$(5.6) \quad L_d(\boldsymbol{\alpha}) = \sum_i \sum_j \alpha_i \alpha_j y_i y_j \mathbf{p}_{ij} - \sum_i^n \alpha_i \mathbf{q}_i + \mathbf{g},$$

where $\mathbf{p}_{ij}, \mathbf{q}_i$ and $\mathbf{g}$ are defined as follows:

$$\begin{cases} \mathbf{p}_{ij} = \left[ \operatorname{tr}\left( \mathbf{w} \mathbf{x}_i^T \mathbf{Z}^T \mathbf{R} \mathbf{Z} \mathbf{x}_j \mathbf{w}^T \right) - \mathbf{w}^T \mathbf{w} \mathbf{x}_i^T \mathbf{Z}^T \mathbf{x}_j \right] \\ \mathbf{q}_i = 1 - y_i b - 2 y_i \mathbf{w}^T \left( \mathbf{X} \mathbf{X}^T \mathbf{Z}^T \right) \mathbf{x}_i \\ -y_i \operatorname{tr}\left[ \left( \mathbf{X} \mathbf{X}^T \mathbf{Z} - 2 \mathbf{X} \mathbf{X}^T \mathbf{Z}^T \mathbf{R} \mathbf{Z} \right) \mathbf{x}_i \mathbf{w}^T \right] \\ -y_i \operatorname{tr}\left[ \mathbf{w} \mathbf{x}_i^T \left( \mathbf{Z}^T \mathbf{X} \mathbf{X}^T - 2 \mathbf{Z}^T \mathbf{R} \mathbf{Z} \mathbf{X} \mathbf{X}^T \right) \right] \\ \mathbf{g} = \operatorname{tr}\left( \mathbf{X} \mathbf{X}^T \right) - 2 \operatorname{tr}\left( \mathbf{X} \mathbf{X}^T \mathbf{Z} \mathbf{X} \mathbf{X}^T \right) \\ -2 \operatorname{tr}\left( \mathbf{X} \mathbf{X}^T \mathbf{Z}^T \mathbf{X} \mathbf{X}^T \right) + 4 \operatorname{tr}\left( \mathbf{X} \mathbf{X}^T \mathbf{Z}^T \mathbf{R} \mathbf{Z} \mathbf{X} \mathbf{X}^T \right) \\ \mathbf{Z} = \frac{1}{2} \left( \mathbf{X} \mathbf{X}^T + \lambda_1 \mathbf{D} + \lambda_2 \mathbf{L} \right)^{-1} \\ \mathbf{R} = \mathbf{X} \mathbf{X}^T + \lambda_1 \mathbf{D} - \lambda_2 \mathbf{L} \end{cases}$$

The detailed steps of the above derivation are available in the supplementary material. Based on Lagrangian duality, $L_d(\boldsymbol{\alpha})$ provides a lower bound for the optimal solution of the original minimization problem w.r.t. $\mathbf{\Phi}$ from Eq. 5.2, as long as the KKT conditions for nonnegativity of the Lagrangian multipliers $\boldsymbol{\gamma}$ and $\boldsymbol{\alpha}$ are satisfied [4]. Hence, to obtain a minimizer for Eq. 5.2, we maximize the dual Lagrangian $L_d(\boldsymbol{\alpha})$, while satisfying the KKT conditions. Note, that $\mathbf{g}$ can be discarded as it does not depend on $\boldsymbol{\alpha}$, leading to the dual optimization:

$$(5.7) \quad \underset{\boldsymbol{\alpha}}{\operatorname{argmax}} \; \frac{1}{2} \boldsymbol{\alpha}^T \mathbf{K} \boldsymbol{\alpha} + \mathbf{q} \boldsymbol{\alpha} \quad \text{s.t } 0 \leq \boldsymbol{\alpha} \leq C^* \mathbf{1},$$

where $\mathbf{K} \in \mathbb{R}^{n \times n}$ is a square matrix with elements $\mathbf{K}_{ij} = 2 y_i y_j \mathbf{p}_{ij}$ and $\mathbf{q} \in \mathbb{R}^n$ is the vector of $-\mathbf{q}_i$ elements. Note, that the added box constraint on $\boldsymbol{\alpha}$ ensures that the non-negativity KKT conditions are satisfied for both $\boldsymbol{\alpha}$ and $\boldsymbol{\gamma}$ (due to Eq. 5.5). The resulting quadratic programming problem is concave (due to Lagrangian duality) and can be efficiently solved by the sequential optimization techniques widely employed in the the SVM literature [17, 21]. The obtained optimal $\boldsymbol{\alpha}$ is employed in Eq. 5.4 to derive the update of $\mathbf{\Phi}$. Note that the diagonal matrix $\mathbf{D}$ depends on $\mathbf{\Phi}$. We update it iteratively, based on the current $\mathbf{\Phi}$ from the previous iteration.

**Algorithm 1** DSL Optimization
───────────────────────────────────────────
**Input:** Training data $(\mathbf{X}, y)$, and parameters $(\lambda_1, \lambda_2, C, \pi)$.
**Output:** The subgraph selection matrix $\mathbf{\Phi}$ and a classifier $(\mathbf{w}, b)$
1: Initialize: $\mathbf{\Phi} \leftarrow \mathbf{I}$
2: **while** $(\mathbf{w}, b)$ and $\mathbf{\Phi}$ have not converged **do**
3: $\quad (\mathbf{w}, b) \leftarrow \operatorname{argmin}_{\mathbf{w}, b} \|\mathbf{w}\|_f + C \sum_i^n \ell\left(y_i, \mathbf{w}^T \hat{\mathbf{x}}_i + b\right)$
4: $\quad$ **while** $\mathbf{\Phi}$ has not converged **do**
5: $\quad\quad \boldsymbol{\alpha} \leftarrow \operatorname{argmax}_{\boldsymbol{\alpha}} \frac{1}{2} \boldsymbol{\alpha}^T \mathbf{K} \boldsymbol{\alpha} + \mathbf{q}\boldsymbol{\alpha}, \text{s.t} 0 \leq \boldsymbol{\alpha} \leq C^* \mathbf{1}$
6: $\quad\quad D_{ii} \leftarrow (2\|\mathbf{\Phi}_i\|_2)^{-1}$ if $\|\mathbf{\Phi}_i\|_2 \neq 0$ or $D_{ii} \leftarrow 0$
7: $\quad\quad \mathbf{\Phi} \leftarrow \frac{(\mathbf{XX}^T + \lambda_1 \mathbf{D} + \lambda_2 \mathbf{L})^{-1}}{2} (\sum_i \alpha_i y_i \mathbf{x}_i \mathbf{w}^T + 2\mathbf{XX}^T)$
8: **return** $\{\mathbf{\Phi}, (\mathbf{w}, b)\}$
───────────────────────────────────────────

**Updates for $(\mathbf{w}, b)$:** When $\mathbf{\Phi}$ is fixed, the optimization of $(\mathbf{w}, b)$ simplifies to the standard linear SVM:

$$(5.8) \quad \operatorname*{argmin}_{\mathbf{w}, b} \|\mathbf{w}\|_f + C \sum_i^n \ell\left(y_i, \mathbf{w}^T \hat{\mathbf{x}}_i + b\right),$$

where minimization of $\|\mathbf{w}\|_f$ ensures maximal margin and the hinge loss penalizes misclassification. It can be solved via quadratic programming (QP) optimization and we rely on efficient solves for this particular quadratic program [14].

The steps of the overall alternating optimization procedure are summarized in Alg. 1. After initialization of the selection matrix and classifier, we repeat the sequential updates until convergence (Steps 2-7). When the subgraph selection $\mathbf{\Phi}$ is fixed, we fit an optimal soft margin SVM for these features in Step 3 and then perform the necessary $\mathbf{\Phi}$ updates (Steps 4-7). The dual Lagrangian is first maximized to obtain an optimal $\boldsymbol{\alpha}$ (Step 5), which is then employed with the current estimate of $D$ (Step 6) in the update for $\mathbf{\Phi}$ (Step 7).

**Complexity analysis.** Due to the enforced sparsity of $\mathbf{\Phi}$, most features loading shrink to zero quickly and as a result the "feature-selected" data matrix $\hat{\mathbf{X}} = \mathbf{\Phi^T X}$ (after projection on $\mathbf{\Phi}$) will be much sparser than the full matrix. If $s_\Phi$ denotes the average number of non-zero elements in $\hat{\mathbf{X}}$, then the complexity of each SVM fit (Step 3) will incur $O(s_\Phi n)$ cost when employing fast sparse solvers [15]. We optimize $\boldsymbol{\alpha}$ by sequential minimal optimization (SMO) which has a cubic complexity $O(n^3)$ in the worse case, but much faster running times have been demonstrated in practice to exhibit between linear and quadratic time costs [22].

| Dataset | $|\mathcal{V}|$ | $|\mathcal{E}|$ | $|\mathcal{DS}|$ | $[\lambda_1, \lambda_2, \pi]$ |
|---|---|---|---|---|
| Synthetic | 100 | 563 | 300 | [ 0.1, 0.3,1] |
| Bike [2] | 142 | 1,723 | 299 | [ 0.5,0.08,1] |
| CCT [25] | 4,665 | 270,571 | 184 | [ 0.1, 0.1,1] |
| ADNI [1, 8] | 6,216 | 683,760 | 173 | [ 0.1,0.01,1] |
| Liver [18, 8] | 7,383 | 251,916 | 123 | [0.05, 0.1,1] |
| Embryo [12, 8] | 1,321 | 5, 227 | 34 | [ 0.1,0.05,1] |

Table 1: Summary statistics of evaluation datasets and the optimal parameters for DSL obtained by cross-validation.

The update of $D$'s diagonal in (Step 6) is linear in the number of non-zero elements in $\mathbf{\Phi}$: $O(s_\Phi)$.

If approached naively the update of $\mathbf{\Phi}$ in Step 7 has a complexity of $O(m^3)$ as it involves an inversion of quadratic in $m$ matrix. However, notice that two of the summands $\mathbf{A} = \mathbf{XX}^T + \lambda_2 \mathbf{L}$ are constants symmetric matrices and so is there sum $\mathbf{A}$, the only varying term in the inversion is the diagonal matrix $\lambda_1 \mathbf{D}$. We can compute the inverse in $O(ms_X)$ time, where $s_X$ is the number of non-zero elements of $\mathbf{X}$ by exploiting this sparse update structure via the Sherman-Morrison formula for sparse inverse updates:

$$(5.9) \quad (\mathbf{A} + \lambda_1 \mathbf{D})^{-1} = \mathbf{A}^{-1} - \frac{\mathbf{A}^{-1}\mathbf{dd}^T\mathbf{A}^{-1}}{1 + \mathbf{d}^T \mathbf{A}^{-1}\mathbf{d}},$$

where $\mathbf{d} = \sqrt{\lambda_1 diag(\mathbf{D})}$ is a column vector, the square root is applied element-wise, and thus $\lambda_1 \mathbf{D} = d^T d$. Note also that the second matrix in $\mathbf{\Phi}$'s update is a linear combination of constant for the inner loop matrices weighted by $\boldsymbol{\alpha}$ plus a globally constant matrix $2\mathbf{XX}^T$ which can be pre-computed once at the cost of $O(min(m^2, s_X^2))$ memory. Assuming this memory cost is paid this matrix can also be computed in $O(ms_X)$.

The total complexity of the method is then $O(t_o s_\Phi n + t_o t_i [n^3 + ms_X + s_\Phi])$, where $t_o$ and $t_i$ are the number of iterations of the outer and inner loops respectively. Assuming constant number of steps to convergence and that $\mathbf{\Phi}$ is sparser than $\mathbf{X}$, the dominating factors in the complexity remain $O(n^3 + ms_X)$, arising from (i) the SMO (Step 5) which as per Platt et al. [22] is at most quadratic as opposed to cubic; and (ii) $\mathbf{\Phi}$ in (Step 7). We employ a standard desktop machine with limited memory for our experiments, and thus, do not perform a high-memory-cost pre-computation, resulting in slightly higher experimental running times which still complete in at most $1h$ for our biggest instances.

## 6 Experimental evaluation

**6.1 Datasets.** We employ both synthetic and real-world datasets for evaluation and summarize their statistics in Tbl. 1.
**Synthetic:** We generate geometric synthetic networks by uniform sampling of node coordinates in a unit square and connecting two nodes if their distance is smaller than a threshold $\tau = 0.2$. We select well-connected subgraphs as the target (ground truth) discriminative subgraph and generate balanced set of in-stances labeled by two global states. Nodes in the target subgraph are randomly assigned values from $[50, 100]$ in positive instances and $[-100, -50]$ in negative counterparts. All remaining nodes are assigned random values from a Gaussian distribution $\mathcal{N}(\mu_i, \sigma^2)$, where $\mu_i$ is the sample mean of the ground truth values in instance $i$ and $\sigma^2$ is a standard deviation which we vary.

**Real-world:** We also employ five real-world datasets. Nodes in the *Bike* [2] are bike rental stations in the Boston city, while edges are connected based on a distance threshold, similar to our synthetic data (see Figure 4(a)). Nodes' feature values correspond to the number of check-outs in a day, while the global states correspond to weekday versus weekends. We employ the last 299 days for this experiment as they span the time when the survice is rolled out to the whole city (initially only the downtown area was covered by the service). The *CCT* dataset contains city cellular HTTP traffic data records [25] for a large city with millions of people. Nodes correspond to stations at which hourly requests are counted (features) and node pairs are once again connected based on a distance threshold. Global labels associated with hourly samples reflect if the sample occurred within workday hours (8am-16pm) or outside this range. The *ADNI* [1] dataset contains fMRI resting state measurements for subjects labeled by AD: suffering Alzheimer's disease and NC: healthy normal controls. The graph structure associates functional links (nodes) with their level of coherence (feature values). Nodes are connected if the corresponding functional links share a brain region. We obtained the ADNI and the two PPI datasets discussed next from the authors of DIPS [8], and thus, have effectively followed the same preparation protocol to enable a fair method comparison. We also employ two gene expression datasets: *Embryo*nic development [12] and *Liver* metastasis [18]. Their network structures are functional protein-protein interaction (PPI) networks [9], while node features correspond to binarized (in Embryo) or continuous (in Liver) gene expression values for healthy and normal subjects (global labels). Prepossessed data was kindly provided by Dang et Al [8].

**6.2 Experimental setup** We evaluate our method's ability to discriminate between global labels, detect ground truth subgraphs, ensure connectivity in the selected subgraphs; and measure overall running time.

**Baselines**: We compare the two flavors of DSL: L1DSL and L2DSL (employing $f = 1$ and $f = 2$ norm for the margin in Eq. 4) to the state-of-the-art method DIPS [8]. We also compare to two recent graph-agnostic feature selectors: FSASL [11] and UDFS [26]. FSASL takes into account an inferred notion of structure among feature angle similarity. UDFS selects discriminative subset from the full set of features and employs the same shrink-enforcing regularized based on th $\ell_{2,1}$ norm. Linear SVM is then employed for prediction on the selected features for the latter two baselines. Our selection of baselines ensures that the state-of-the-art graph-aware method is considered, as well as graph-agnostic alternatives which enforce sparsity and margin maximization similar to DSL, which constitute design advantages lacking in DIPS [8].

**Metrics**: When ground truth (GT) desired feature subgraphs are available, we calculate the area under the ROC (AUC) for recovering GT nodes in the selection. We quantify the testing prediction accuracy based on selected subgraphs (feature subsets) in 5-fold stratified cross-validation (CV). We also measure the "community" structure of selected subgraphs in terms of the conductance $\phi$ of their induced subgraphs within the summary network structure $S$.

**Implementation**: Our methods are implemented in Matlab 2017b and all reported running times are for single-core (non-parallel) execution on an Intel(R) Xeon(R) Gold 6138 CPU @ 2.00GHz processor in a Dell PowerEdge system.

**6.3 Classification accuracy** We first evaluate the ability of competing methods' selected features to discriminate between global states in cross-validation on all real-world datasets (Fig. 2). For this experiments we fix the number of selected features for each of the competing techniques (we varying between 1% and 2% of all features in the respective datasets) and train a linear SVM (C=1) on only selected features. L1DSL consistently outperforms all alternatives for varying number of features with the gap in performance from the best baseline being highest for small number of selected features. The superior performance is due to the simultaneous sparse and consistent selection of discriminative connected subgraphs. Each baseline enforces only a subset of all those requirements: (i) DIPS employs a non-sparse discriminative subspace learning, which is then in an independently thresholded and smoothed against the graph structure; (ii) FSALSL and UDFS do not take advantage of the graph structure.

In the *Bike* dataset (Fig. 2a), both DSL variants exhibit the largest improvement (15% higher accuracy) with as little as 10 selected features. More importantly DSL methods reach very close to the saturation accuracy of 95% with as little as 5 features. While the gap from DIPS closes, the latter continues to underperform DSL with higher number of features. L1DSL consistently outperforms alternatives by 10% for subgraph selection sizes between 40 and 100 on the *CCT* dataset Fig. 2b. On this dataset L2DSL is slightly worse for smaller number of features and all methods' performance degrades significantly when restricted to less than 40 features. All alternatives do not exceed 73% accuracy and reach their peaks they employ as many as 90 (DIPS), 70 (FSASL) and 60 (UDFS) features, while L1DSL reaches its peak with 40-node subgraphs (1% of all features in CCT). Both the CCT and Bike are spatial datasets in which local connectivity matters in selecting connected discriminative subgraph patterns, thus, resulting in the biggest advantage for DSL methods.

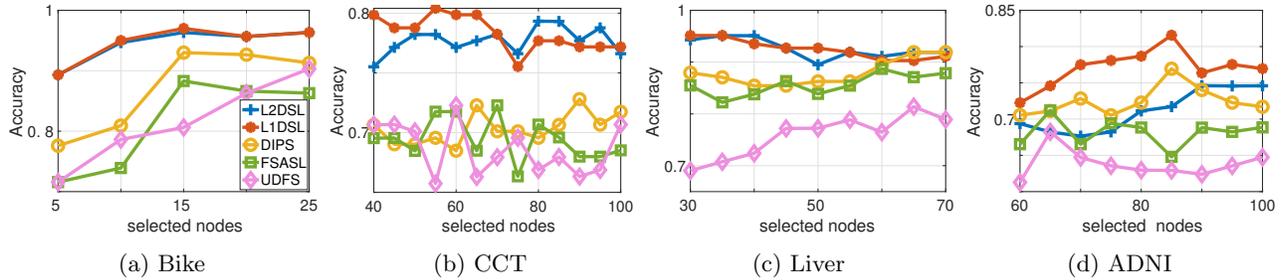

(a) Bike  (b) CCT  (c) Liver  (d) ADNI

Figure 2: Comparison of prediction accuracy in 5-fold cross-validation on real datasets for increasing number of selected features by each of the competing techniques.

The advantage of our methods in *Liver* (Fig. 2c) is also most evident when restricted to small number of features (both DSL variants are indistinguishable on this data). DSL's performance peaks at 95%, employing 30-node subgraphs, while DIPS reaches its highest 93% with more than twice the number of features. The other baselines also require higher number of features to reach their maximal accuracy. L1DSL dominates alternatives in *ADNI* reaching accuracy of 82% with 85 nodes, while the best accuracy of DIPS is 77% with also 85 nodes. UDFS and FSASL perform significantly worse on this data. L2DSL does not perform on par with L1SVM on ADNI, which could be explained by its higher propensity to consider more and redundant features to maximize the margin as it enforces less shrinkage via an L2 as opposed to L1 norm on $w$. ADNI and Liver are both complex networks (unlike Bike and CCT), featuring high node degrees and potentially some edges which are "less aligned" with the underlying process which determines the global network state. The optimal graph smoothness regularizer weights $\lambda_2$ for these datasets are also lower (see Tbl. 1), corroborating the hypothesis of comparatively lower importance of the network structure.

## 7 Quality of feature selection

When ground truth features of interest are available, we compare the techniques by their ability to recover these GT features in their selected subgraphs. We first compare L1DSL (L2DSL's performance is indistinguishable from L1DSL's here) and DIPS on synthetic data and increasing variance $\sigma^2$ applied to non-GT nodes' values (Fig. 1). Here we restrict both methods to report 15 features and provide a visual comparison of the selected features by DSL (squares) and DIPS (crosses) and the GT nodes (large round circles) in Fig. 1a ($\sigma^2 = 40$). DSL matches the well-connected ground truth exactly, while DIPS selects only a subset of those nodes and also include noisy singleton nodes. We have selected the optimal graph smoothness regularizer for DIPS, so enforcing more smoothness for this method leads to strictly poorer AUC of recovering the GT. In other words DIPS is more sensitive to noise as in its first step (subspace projection) it considers all features to compute its cross-sample similarity graphs. As we increase the variance of values in non-GT nodes, the GT-detection AUC decreases drastically for DIPS, while remaining more stable for L1DSL, opening a performance gap of 24% at $\sigma^2 = 10^2$ (Fig. 1b).

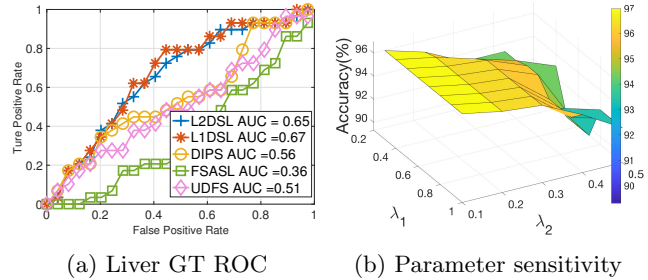

(a) Liver GT ROC  (b) Parameter sensitivity

Figure 3: (a) ROC for selecting ground truth genes (from [18]) in Liver. (b) Parameter sensitivity on the Bike dataset varying $\lambda_1$ and $\lambda_2$.

We also quantify the feature selection quality for the *Liver* dataset for which we use the GT genes associated with the disease reported in the original paper [18] Fig. 3a. In this experiment we plot the ROC curves for the competing techniques. At small FPRs, DSL methods perform similar to DIPS and UDFS, however, the TPRs of L1DSL and L2DSL grow at a faster rate than that of alternatives. At FPR=0.5, the TPR of all alternatives does not exceed 0.5, while both DSL methods achieve a TPR of 0.8. It is important to note that the GT set of genes is likely incomplete resulting in limited TPR growth at low FPR regimes. However, the newly predicted genes by DSL are likely going to provide good targets for additional genes associated with the diseases as their selection optimizes both smoothness w.r.t. the PPI structure (guilt by association) and their discriminative power for the global state.

|          |      | Bike | CCT  | ADNI | Liver | Embryo |
|----------|------|------|------|------|-------|--------|
| **Accuracy** | DSL  | 0.89 | 0.81 | 0.78 | 0.95  | 0.88   |
|          | DIPS | 0.73 | 0.69 | 0.71 | 0.88  | 0.80   |
| $\phi$   | DSL  | 0.9  | 0.12 | 0.91 | 0.82  | 0.90   |
|          | DIPS | 0.97 | 0.98 | 0.83 | 0.99  | 0.97   |
| **Time** | DSL  | 2s   | 28m  | 45m  | 61m   | 78s    |
|          | DIPS | 1s   | 9m   | 4m   | 5m    | 15s    |

Table 2: Comparison of DSL and DIPS on real-world datasets in terms of accuracy, conductance $\phi$ of the discovered subgraphs and running time. Methods are restricted to select the number of features resulting in the larges accuracy gap in each dataset.

**7.1 Parameter sensitivity** While our model requires three parameters: $\lambda_1$, $\lambda_2$ and $\pi$, it is not very sensitive to their values. This is demonstrated by the relatively stable optimal parameters selected across datasets by cross-validation (last column of Tbl. 1). Particularly, the optimal weight of the SVM margin maximization $\pi$ is always one (and the performance is similar for a range of values). There is more variability in the optimal selections of $\lambda_1$ and $\lambda_2$ across datasets (typically small values between 0.01 and 0.1), however, the resulting accuracy is stable across wide ranges of those values within a dataset. We demonstrate this behavior by plotting the accuracy as a function of $\lambda_1$ and $\lambda_2$ for the Bike dataset in Fig. 3(b). In this dataset the accuracy remains 4% of the optimal accuracy for wide ranges of the parameters. We observe similar trends in other datasets and almost no variation when taking $\pi$ between [1, 5].

**7.2 Quality, connectivity and running time.** Table 2 offers a comprehensive comparison of DIPS and DSL on all dataset, where accuracy is presented alongside with running time and conductance of the selected subgraphs. The best accuracy gap separation is at least 7% and reaches up to 16% difference on the Bike data. The selected subgraphs by DSL, not only have higher discriminative power, but are also better connected (lower conductance values signifies lower cut to volume ratio for selected node). One exception to this trend is the ADNI dataset in which the conductance of DSL's solution is slightly slower. Note that the structure of this dataset is very regular: $S$ is the dual graph of a fully-connected coherence network among graph regions, and hence since there are no good cuts in this graph the conductance community measure is less informative. Our methods' superior quality come at with the cost of running time than that of DIPS. The main reason is our joint connected discriminative subgraph selection, which requires more computations than DIPS' two-step independent, though less accurate, optimization. Our methods' implementation complete in at most an hour on the biggest evaluation datasets. It is important to note that our current DSL implementation does not employ the optimal pre-computation of large static matrices (as discussed in Sec. 5), hence, its running times could be improved at the cost of higher memory footprint.

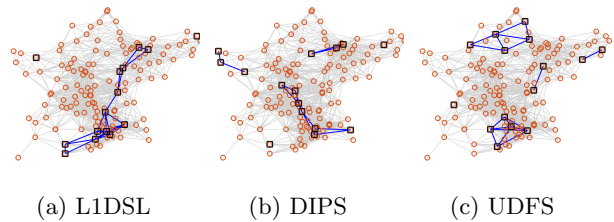

(a) L1DSL  (b) DIPS  (c) UDFS

Figure 4: Visualization on Boston Bike Trips Data and the results by different methods.

**7.3 Discriminative subgraphs in bike commutes** We visualize the selected subgraphs in the Bike data in Fig. 4, where nodes are plotted according to the geographical coordinates of bike rental stations in the city of Boston. Consistent with the analysis on synthetic, DIPS and UDFS tend to select more disconnected subgraph which collectively have lower discriminative power for weekday versus weekend rental patterns. Interestingly, L1DSL selects a long NE-SW "corridor" which passes through downtown as the most discriminative, leading to an intuitive interpretation that this corridors rental pattern might discriminate well between weekday commutes to the downtown areas from the periphery and the likely lack thereof during weekends.

## 8 Conclusion

We presented a novel optimization framework for the problems of global network state prediction and feature selection, called DSL. It enforces simultaneously the natural requirements for sparsity, connectivity in the network structure, and discriminative power of selected subgraph solutions. We demonstrated DSL's superior quality when employed on both synthetic and real-world problem instances and in comparison to state-of-the-art baselines. Our method was able to recover ground truth subgraphs in synthetic and gene expression datasets with consistently better accuracy than competitors. The subgraphs learned by DSL enabled between 7% and 16% improvement of cross-validation classification accuracy compared to the closest baseline. In addition, we demonstrated the interpretability and applicability DSL's solutions by uncovering known target genes involved in liver metastasis and by uncovering intuitive commute subgraph patterns in transportation networks capable of distinguishing between workday and weekend activity.

## Acknowledgements

We would like to thank X-H. Dang and H. You, authors of DIPS [8], for kindly sharing evaluation datasets and the implementation of their algorithm, as well as for several informative discussions.

## References

[1] ADNI project. http://www.adni-info.org/.

[2] Hubway data visualization challenge:. http://hubwaydatachallenge.org.

[3] P. Bogdanov, N. Dereli, X.-H. Dang, D. S. Bassett, N. F. Wymbs, S. T. Grafton, and A. K. Singh. Learning about learning: Mining human brain sub-network biomarkers from fmri data. *PLoS One*, 2017.

[4] S. Boyd and L. Vandenberghe. *Convex optimization*. Cambridge university press, 2004.

[5] G. Chandrashekar and F. Sahin. A survey on feature selection methods. *Computers & Electrical Engineering*, 40(1):16–28, 2014.

[6] C. Cortes and V. Vapnik. Support-vector networks. *Mach. Learn.*, 20(3):273–297, Sept. 1995.

[7] X. H. Dang, A. K. Singh, P. Bogdanov, H. You, and B. Hsu. Discriminative subnetworks with regularized spectral learning for global-state network data. In *Proceedings of the European Conference on Machine Learning and Knowledge Discovery in Databases - Volume 8724*, ECML PKDD 2014, pages 290–306, Berlin, Heidelberg, 2014. Springer-Verlag.

[8] X. H. Dang, H. You, P. Bogdanov, and A. K. Singh. Learning predictive substructures with regularization for network data. In *2015 IEEE International Conference on Data Mining*, pages 81–90, Nov 2015.

[9] R. Dannenfelser, N. R. Clark, and A. Ma'ayan. Genes2fans: connecting genes through functional association networks. *BMC Bioinformatics*, 13(1):156, Jul 2012.

[10] M. Das Gupta and J. Xiao. Non-negative matrix factorization as a feature selection tool for maximum margin classifiers. In *Proceedings of the 2011 IEEE Conference on Computer Vision and Pattern Recognition*, CVPR '11, pages 2841–2848, Washington, DC, USA, 2011. IEEE Computer Society.

[11] L. Du and Y.-D. Shen. Unsupervised feature selection with adaptive structure learning. In *Proceedings of the 21th ACM SIGKDD International Conference on Knowledge Discovery and Data Mining*, KDD '15, pages 209–218, New York, NY, USA, 2015. ACM.

[12] J. Dutkowski and T. Ideker. Protein networks as logic functions in development and cancer. *PLOS Computational Biology*, 7(9):1–11, 09 2011.

[13] E. Elhamifar and R. Vidal. Sparse subspace clustering: Algorithm, theory, and applications. *IEEE Trans. Pattern Anal. Mach. Intell.*, 35(11):2765–2781, Nov. 2013.

[14] R.-E. Fan, K.-W. Chang, C.-J. Hsieh, X.-R. Wang, and C.-J. Lin. LIBLINEAR: A library for large linear classification. *Journal of Machine Learning Research*, 9:1871–1874, 2008.

[15] T. Joachims. Training linear svms in linear time. In *Proceedings of the 12th ACM SIGKDD International Conference on Knowledge Discovery and Data Mining*, KDD '06, pages 217–226, New York, NY, USA, 2006. ACM.

[16] M. G. Kann. Protein interactions and disease: computational approaches to uncover the etiology of diseases. *Briefings in Bioinformatics*, 8(5):333–346, 2007.

[17] S. Keerthi and E. Gilbert. Convergence of a generalized smo algorithm for svm classifier design. *Machine Learning*, 46(1):351–360, Jan 2002.

[18] D. H. Ki, H.-C. Jeung, C. Park, S. Hee Kang, G. Youn Lee, W. S. Lee, N. Kyu Kim, H. Chung, and S. Young Rha. Whole genome analysis for liver metastasis gene signatures in colorectal cancer. 121:2005–12, 11 2007.

[19] G. Liu, Z. Lin, and Y. Yu. Robust subspace segmentation by low-rank representation. In *Proceedings of the 27th International Conference on International Conference on Machine Learning*, ICML'10, pages 663–670, USA, 2010. Omnipress.

[20] Y. Nesterov. A method of solving a convex programming problem with convergence rate O(1/sqr(k)). *Soviet Mathematics Doklady*, 27:372–376, 1983.

[21] J. Platt. Sequential minimal optimization: A fast algorithm for training support vector machines. Technical report, April 1998.

[22] J. Platt. Sequential minimal optimization: A fast algorithm for training support vector machines. 1998.

[23] S. Ranu, M. Hoang, and A. Singh. Mining discriminative subgraphs from global-state networks. In *Proceedings of the 19th ACM SIGKDD International Conference on Knowledge Discovery and Data Mining*, KDD '13, pages 509–517, New York, NY, USA, 2013. ACM.

[24] N. Srebro, J. D. M. Rennie, and T. S. Jaakkola. Maximum-margin matrix factorization. In *Proceedings of the 17th International Conference on Neural Information Processing Systems*, NIPS'04, pages 1329–1336, Cambridge, MA, USA, 2004. MIT Press.

[25] S. Q. W. H. K. J. Xiaming Chen, Yaohui Jin. Analyzing and modeling spatio-temporal dependence of cellular traffic at city scale. In *Communications (ICC), 2015 IEEE International Conference on*, 2015.

[26] Y. Yang, H. T. Shen, Z. Ma, Z. Huang, and X. Zhou. L2,1-norm regularized discriminative feature selection for unsupervised learning. In *Proceedings of the Twenty-Second International Joint Conference on Artificial Intelligence - Volume Volume Two*, IJCAI'11, pages 1589–1594. AAAI Press, 2011.

[27] J. Zhu, S. Rosset, R. Tibshirani, and T. J. Hastie. 1-norm support vector machines. In S. Thrun, L. K. Saul, and B. Schölkopf, editors, *Advances in Neural Information Processing Systems 16*, pages 49–56. MIT Press, 2004.